\documentclass{article}

\usepackage[nonatbib, final]{neurips_2021}

\usepackage[utf8]{inputenc} 
\usepackage[T1]{fontenc}    
\usepackage{hyperref}       
\usepackage{url}            
\usepackage{booktabs}       
\usepackage{amsfonts}       
\usepackage{amsmath}        
\usepackage{nicefrac}       
\usepackage{microtype}      
\usepackage{xcolor}         
\usepackage{mathtools}
\usepackage{float}

\usepackage{caption}
\usepackage{subcaption}

\usepackage{tikz}
\usetikzlibrary{positioning, arrows}

\usepackage[ruled,vlined]{algorithm2e}
\usepackage{bm}

\usepackage[
backend=bibtex,
maxbibnames=30,
sorting=none,
date=year,
doi=false,
isbn=false,
url=false
]{biblatex}
\newcommand{\E}{\ensuremath{\mathbb{E}}}				
\newcommand{\x}{\ensuremath{\bm{x}}}					
\newcommand{\xhat}{\ensuremath{\bm{\hat{x}}}}		    
\newcommand{\X}{\ensuremath{\bm{X}}}					
\newcommand{\z}{\ensuremath{\bm{z}}}					

\newcommand{\enc}{\ensuremath{enc_\phi}}				
\newcommand{\dec}{\ensuremath{dec_\theta}}			
\newcommand{\cox}{\ensuremath{cox_\psi}}	  			

\DeclarePairedDelimiterX{\infdivx}[2]{(}{)}{%
	#1\;\delimsize\|\;#2%
}
\newcommand{\kl}[2]{\ensuremath{\mathbb{KL}\infdivx{#1}{#2}}} 
\newcommand{\loss}[1]{\ensuremath{\mathcal{L}_{#1}}}		

\newcommand{\pinf}{\ensuremath{q_\phi(\bm z \mid \bm x)}}			

\newcommand{\lossbvae}{\ensuremath{\loss{\beta {\scriptscriptstyle - VAE}}}}	

\newcommand{\eqendp}{\,\text{.}} 

\title{Survival-oriented embeddings for improving accessibility to complex data structures}

\author{%
  Tobias Weber\\
  Department of Statistics\\
  LMU Munich\\
  \texttt{tobias.weber@stat.uni-muenchen.de} \\
   \And
  Michael Ingrisch\\
  Department of  Radiology\\
  LMU Munich\\
  \texttt{michael.ingrisch@med.uni-muenchen.de} \\
  \And
  Matthias Fabritius\\
  Department of  Radiology\\
  LMU Munich\\
  \texttt{matthias.fabritius@med.uni-muenchen.de} \\
  \And
  Bernd Bischl\\
  Department of Statistics\\
  LMU Munich\\
  \texttt{bernd.bischl@stat.uni-muenchen.de} \\
   \And
  David R\"ugamer\\
  Department of Statistics\\
  LMU Munich\\
  \texttt{david.ruegamer@stat.uni-muenchen.de} \\
}

\begin{document}

\maketitle

\begin{abstract}

Deep learning excels in the analysis of unstructured data and recent advancements allow to extend these techniques to survival analysis.
In the context of clinical radiology, this enables, e.g., to relate unstructured volumetric images to a risk score or a prognosis of life expectancy and support clinical decision making.
Medical applications are, however, associated with high criticality and consequently, neither medical personnel nor patients do usually accept black box models as reason or basis for decisions. Apart from averseness to new technologies, this is due to missing interpretability, transparency and accountability of many machine learning methods.
We propose a hazard-regularized variational autoencoder that supports straightforward interpretation of deep neural architectures in the context of survival analysis, a field highly relevant in healthcare.
We apply the proposed approach to abdominal CT scans of patients with liver tumors and their corresponding survival times.  
\end{abstract}



\section{Introduction}

Recent developments in survival analysis (SA) have shown a rising interest in the modeling of unstructured data types using deep learning. In medical applications these data types often constitute (volumetric) CT or MRI scans, but other data types such as audio or video signals are also not uncommon.
While a majority of deep learning approaches in SA either focus on tabular data \parencite{Katzman2016,Luck2017, Lee2018, Lee2019, Lee2020, Groha2020, bender2021general} or directly apply convolutional neural networks to predict survival based on image data \cite{Zhang2019, Haarburger2018, li2020using}, more recent methods such as \parencite{Kopper2020a, Ruegamer2021} combine tabular and unstructured information.

%
%
%
\paragraph{From machine learning research to clinical practice} 

Given these methodological advancements and the rising interest in artificial intelligence as well as in digital health systems, machine learning models might be used to predict risk scores or patient survival in the near future and play an important role in clinical decision making. This is, however, an application with high criticality and consequently high requirements for transparency and explainability \parencite{DataEthic2019}. Both are considered a must, as model predictions imply decisions such as appropriate therapy measures or a patient's risk assessment. 

Radiology with its comparatively large amounts of unstructured imaging data plays a pivotal role in the digitalization of clinical medicine.
As a consequence, numerous efforts have been undertaken to link unstructured imaging data to well defined clinical endpoints such as diagnosis, i.e., image classification, and prognosis, i.e., survival analysis. This is often referred to as radiomics.
Since overall survival is an important endpoint in clinical trials, radiomics has also been extended to survival analysis.
As in many other domains, deep learning (DL) is a promising candidate for radiological image analysis, but is currently limited in its application.
This is not only due to limited sample sizes, but also because of the black-box character of typical DL models.
Providing interpretability, explainability and transparency in radiological applications is thus the next important step towards a successful implementation of modern analysis tools in clinical radiology.
%
%
%

\paragraph{Interpretablity methods}

Methods for interpretability often analyze and visualize activations in classification or regression tasks to explain the model's reasoning.
For example in \cite{Selvaraju2016, Chattopadhyay2017} gradient-weighted activations are utilized to create a mapping similar to a heatmap that highlights decisive areas in the original image.
This is also a useful tool in radiological applications where, e.g., an activation centered in the lung region of a chest X-ray increases trust in a model's prediction for the presence of pneumonia.

\begin{figure}[htbp]
    \centering
    \includegraphics[width=\textwidth]{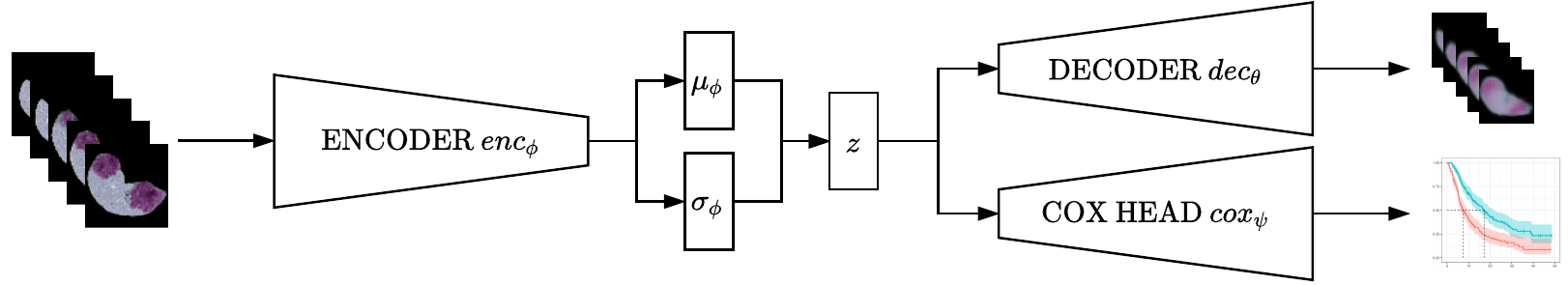}
    \caption{CoxVAE multi-task architecture. The encoder \enc\ estimates the low-dimensional latent feature distribution $\z | \x \sim \mathcal{N}(\bm \mu_\phi, \bm \sigma^2_\phi \bm I)$ for an arbitrary input \x, e.g., a CT image. Based on the learned embedding $\z$, the decoder reconstructs the image \xhat\ while \cox\ predicts the patient's survival by estimating the log-hazard rate based on the corresponding latent sample.}
    \label{fig:coxvae-archi}
\end{figure}

However, to the best of our knowledge, no dedicated tool for the interpretation of models in deep survival learning exists as of now.
To close this gap, we propose \emph{CoxVAE} (Fig.~\ref{fig:coxvae-archi}), a Cox-regularized variational autoencoder (VAE; \parencite{Kingma2014}) that allows for meaningful compression of unstructured data and a straightforward interpretation of the model's reasoning.
Similar to \cite{Bello2019}, CoxVAE utilizes survival label information to produce a survival-optimized embedding that allows to directly assess the importance of the obtained latent features and meaningful processing for subsequent survival downstream applications.

\section{Survival-oriented embeddings in medical imaging}


The VAE is a common and established choice for encoding data. Let \enc\ be the probabilistic encoder with parameters $\phi$ that predicts a latent vector of variables \z\ with distribution $\z \mid \x \sim \mathcal{N}(\bm \mu_\phi, \bm \sigma^2_\phi \bm I)$ using a sample $\x \in \mathcal{X}$ from a sample space $\mathcal{X}$. Given $\bm{z}$, the decoder \dec\ with parameters $\theta$ creates a reconstruction \xhat\ of \x. The VAE can be trained by minimizing the negative evidence lower bound (ELBO)
\begin{equation} \label{eq:theo-vae-beta}
	\lossbvae(\theta, \phi; \x) =  - \E_{\z \sim \pinf} \left[ \log p_\theta(\x \mid \z) \right] + \beta \kl{q_\phi(\bm z \mid \bm x)}{p (\z)} \eqendp
\end{equation}
The ELBO in \eqref{eq:theo-vae-beta} represents a trade-off between good reconstruction, i.e., maximizing $\log p_\theta(\x | \z)$, and minimizing the KL-divergence $\mathbb{KL}$ between the variational density $q_\phi(\bm z | \bm x)$ and the Gaussian prior $p (\z)$, controlled by the parameter $\beta$. This type of autoencoder is also widely known as $\beta$-VAE \parencite{alemi2019deep, higgins2016beta, Burgess2018}.


\paragraph{Cox survival objective}

The reconstruction of the ($\beta$-)VAE induces a latent space that comprises as much information as possible about the data \x, but does not necessarily have an intuitive interpretation. This is, however, a indispensable property when reusing these latent representations in other models or downstream tasks. Especially for medical applications, an interpretable latent space is crucial and helps clinicians in understanding image features such as detected tumor tissue.
To address this need, we propose a CoxVAE framework that extends the base VAE architecture with an additional network head \cox\ with parameters $\psi$. The network head \cox\ itself is a Cox PH model \cite{Cox1972} that estimates the log-hazard rate $r = \cox(\z) \in \mathbb{R}$. More specifically, for a dataset $\bm X$ of $n$ observations $(t_i, \delta_i, \bm{x}_i)_{i=1,\ldots,n}$ with event times $t_i$, boolean censoring indicator $$\delta_i = \begin{cases} 0 & \text{if censored} \\ 1 & \text{else}, \end{cases}$$ and images $\bm{x}_i$, latent variables $\z_i \sim q_\phi(\bm z \mid \bm x_i)$ are first generated by the VAE. Based on the negative partial Cox log-likelihood (c.f. \cite{Katzman2016})
\begin{equation} \label{eq:meth-coxvae-coxloss}
	\mathcal{L}{\scriptscriptstyle Cox}(\phi,\psi; \X) = - \frac{1}{\sum_{i=1}^{n} \delta_i} \sum_{i=1}^{n} \delta_i \left( r_i - \log \sum\limits_{j: t_j \geq t_i} \exp(r_j)  \right)
\end{equation}
and the predicted log-hazard rates $r_i = \cox(\z_i)$, the parameters $\psi$ can be optimized in a second step.  


\paragraph{Combined objectives}

Instead of a two-step procedure, we combine the two objectives \eqref{eq:theo-vae-beta} and \eqref{eq:meth-coxvae-coxloss} and jointly learn the \enc, \dec, and \cox\ (schematically depicted in Figure~\ref{fig:coxvae-archi}). The introduction of a supervised survival loss forces the learned latent space to incorporate the information of the survival labels, which in turn allows for a meaningful compression and better interpretability. As the CoxVAE constitutes a multi-task architecture, we propose to control the balance of both losses via a parameter $\tau \in [0, 1]$ in the final objective function:
\begin{equation} \label{eq:meth-total-loss}
	\mathcal{L}_(\phi, \theta, \psi; \X) = \frac{\tau}{n} \sum_{i=1}^n \left( \lossbvae(\phi, \theta; \x_i) \right) + (1 - \tau) \mathcal{L}_{\scriptscriptstyle Cox}(\phi,\psi; \X) \eqendp
\end{equation}
A large value for $\tau$ implies a focus on reconstruction, whereas a small value leads to an embedding that strongly influenced by the survival times.


\section{Survival of cancer patients with liver metastases}

We trained the proposed CoxVAE on anonymized contrast-enhanced computed tomography scans of 492 patients with liver metastases.  For each patient, the dataset contains an abdominal CT scan and the time until death in days. The overall censoring rate is at 17\%.

\paragraph{Modeling approach}

To obtain segmentation masks of the liver, we use the nnU-Net \parencite{Isensee2018a}. The CT scan is then downscaled to a resolution of $64 \times 64 \times 64$ voxels. For \enc\ and \dec\ we choose neural networks with four residual blocks \parencite{He2016} and a latent space of $\mathbb{R}^8$. While the choice of \cox\ can be an arbitrary Cox PH model, we found that a linear predictor without bias term yields good results while being inherently interpretable. 
Another important choice for the model's performance is the correct balance between the complexity of the two heads \dec\ and \cox. As \cox\ in comparison to \dec\ only needs a fraction of updates to converge due to the considerably smaller network, we employ two optimizers (Adam; \parencite{Kingma2015}) instead of only one joint optimizer and define different learning rates for both network parts ($1\text{e-}4$ for $\theta, \phi$ and  $1\text{e-}5$ for $\psi$). Training is conducted for 16,000 batch updates with a batch size of 16.

\section{Results}

\begin{figure}[t]
     \centering
    \includegraphics[width=\textwidth]{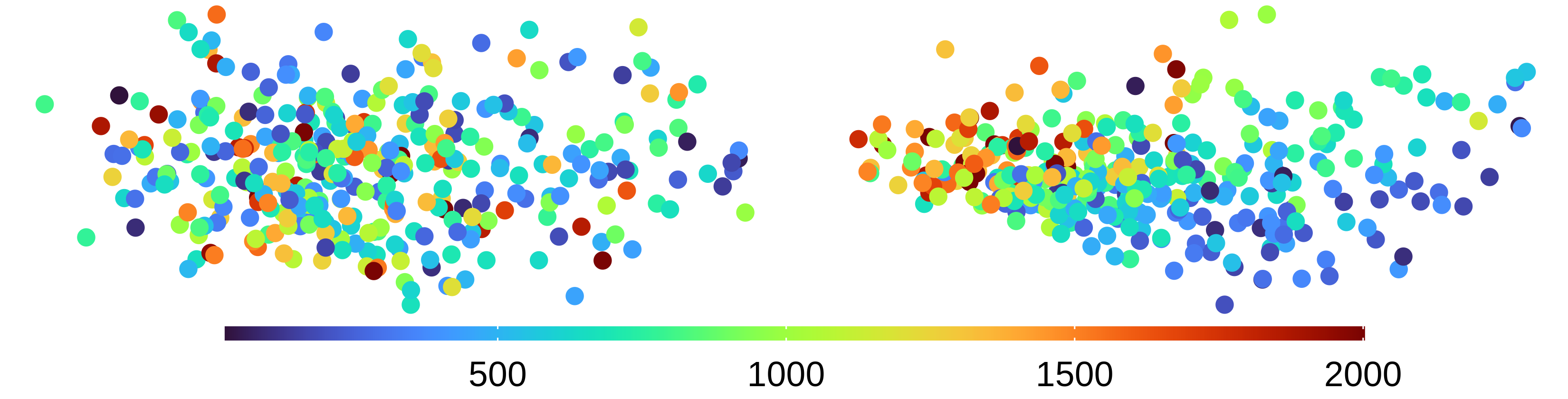}
    \caption{Comparison of the first two PCA components of a VAE (\textbf{left}) and a CoxVAE (\textbf{right}) embedding for CT imaging data. The color indicates survival time in days. The CoxVAE embedding reflects the learned survival information, whereas the unsupervised VAE shows no apparent grouping.}
    \label{fig:sirflox-emb}
\end{figure}

The main goal of our analysis is to compress CT scans and create an embedding optimized for further survival downstream tasks and straightforward interpretability.
While the VAE as unsupervised model has to rely on the visual structures in the supplied images, the CoxVAE is able to incorporate more distinct label information through \cox\ in a supervised manner.
The results is a completely restructured latent space as depicted in Figure~\ref{fig:sirflox-emb}.
For the vanilla VAE, a PCA dimension reduction of the latent space shows that the main variation in the data does not involve grouping or clusters with similar survival times. Instead, data points are mainly clustered based on a visual keys (e.g., the shape and size of the liver).
In contrast, the CoxVAE embedding shows a distinct ordering of the PCA's first component along with decreasing survival times.
The focus on visual aspects of the images is a common shortcoming of the VAE, potentially only focusing on low frequency features such as the coarse shape of the liver, whereas high frequency features, e.g., small tumor patches, are often neglected.
By employing the additional survival-head, we solve this issue as features that are visually less relevant but crucial for survival receive a greater weight in the total objective \eqref{eq:meth-total-loss} (and vice versa).

\paragraph{Inspecting latent dimensions}

While \cox\ can be chosen to be any Cox PH-based model (e.g., a DeepSurv architecture \cite{Katzman2016}), using a single linear layer network allows for further insights into the model's reasoning and its latent space. 
\begin{figure}[htbp]
     \centering
    \includegraphics[width=0.95\textwidth]{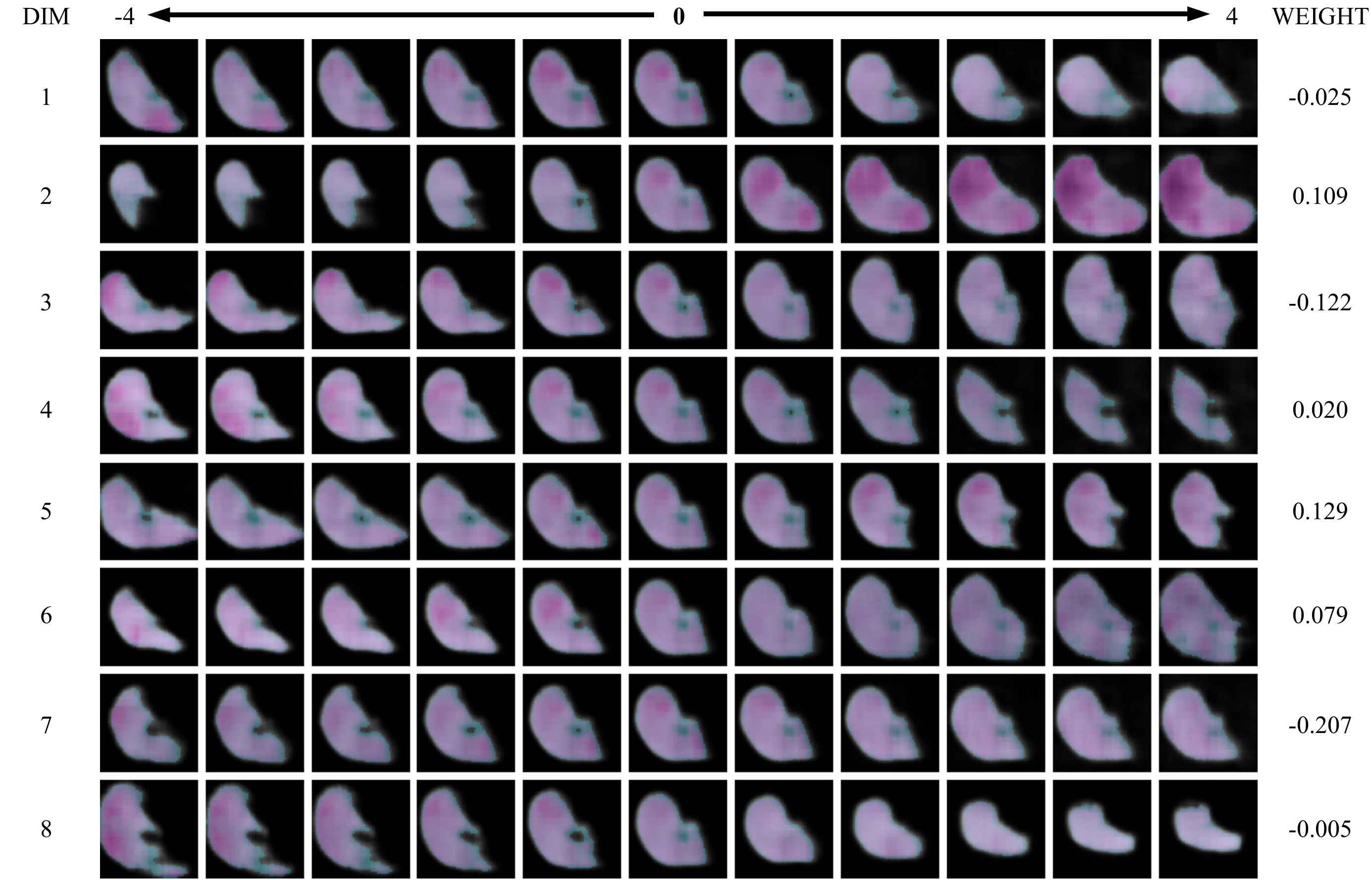}
    \caption{Traversing across all latent 8 latent dimensions in CoxVAE with linear \cox. The images show the reconstruction of different \z\ of CTs with liver metastases. For reasons of clarity, we only visualize the middle slice of the 3D image. Tumor patches are highlighted with a red overlay. The base \z\ (\textbf{middle column}) is a zero vector. The \textbf{columns} model the effect of changing the value in the respective dimensions (\textbf{rows}) in the range -4 to 4. On the \textbf{right} side the weight of \cox\ is displayed, that is associated with each dimension. Latent dimensions with a \cox\ weight close to zero are deemed as rather irrelevant for estimating survival. Large positive weights and large value in \z\ correspond with a large hazard rate. On the contrary, large negative values result in a high hazard rate when the \cox\ weight is large negative.}
    \label{fig:sirflox-dim-walk}
\end{figure}
Being a linear model, a one unit change in the $i$-th dimension $\psi_i$ of $\bm{\psi}$ changes the resulting hazard ratio by a factor of $\exp(\psi_i)$. 
As the latent dimensions are mostly disentangled due to the VAE prior assumption, each dimension can thus be assessed through their importance for the survival outcome and changes in these specific latent features can be represented in $\mathcal{X}$ by utilizing \dec.
Figure~\ref{fig:sirflox-dim-walk} exemplarily depicts this for the CT liver scans with tumor metastases and an 8-dimensional latent space.
In this example, dimension 2 of the latent space has an assigned weight of 0.109 (on the log-scale), which translates to an 11.5\% increase of the hazard rate per unit change in the latent dimension.
As a result, the reconstructed images show a distinct increase and spread of tumor patches along this axis when increasing $\phi_2$ (or vice versa, decreasing $\phi_2$ results in disappearing tumor patches).
%
%
%
%


\paragraph{Impact of $\tau$-parameter}

\begin{figure}[htbp]
     \centering
    \includegraphics[width=\textwidth]{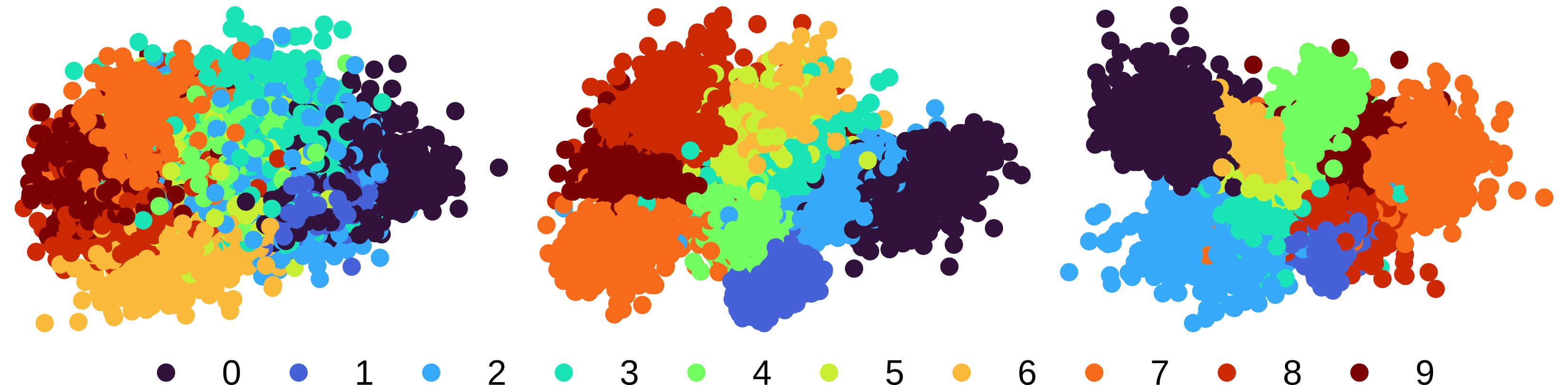}
    \caption{First two PCA-components for encoded surv-MNIST with a $\tau$ value of 0.01 (\textbf{left}), 0.5 (\textbf{middle}) and 0.99 (\textbf{right}). With increasing $\tau$ the embedding forms more distinct clusters. The main source of variation changes from the explanation of survival time to structural similarities.
    }
    \label{fig:mnist-tau}
\end{figure}

An important parameter in our proposed model is $\tau$, which controls the amount of reconstruction as well as the focus on the survival task. Extreme values of $\tau$ either allow to solely focus on reconstruction or on good survival prediction. In order to investigate its behaviour, we conducted further simulation studies with knowledge of the ground truth based on synthetic MNIST images. 
Following \cite{Gensheimer2019} the synthetic dataset assigns low hazard rates to low digits and vice versa.
Figure~\ref{fig:mnist-tau} depicts the evolution of the latent space along different $\tau$ values. As expected, small values of $\tau$ imply a strong focus on the survival task, visible by the learned natural order of the digits along the first component of a PCA. An increase of $\tau$ results in more distinct digit clusters and disentanglement. For large $\tau$ values, the neighborhood of digits shifts to visual similarities, e.g., the digit \texttt{6} is now close to digit \texttt{0}, whereas \texttt{0} and \texttt{1} are far away from each other. These results confirm our theoretical presumptions and underline the importance of this parameter. In practice, we recommend choosing $\tau$ as small as possible (with as much emphasis on the survival task as possible), while ensuring reconstructed images to still look meaningful and interpretable for clinicians.
%


\section{Practical considerations}




An almost omnipresent challenge for the application of survival analysis to radiological imaging data is the limited data size.
On the one hand, clinical trials typically only collect tens or hundreds of patients.  On the other hand, deep learning-based approaches require a multiple of training data for yielding well-generalizing results with low bias.
Although the  dataset used in this demonstration is considered large in radiology, fitting a generative deep learning architecture for high-dimensional image data is still challenging. 
In many applications, it is also not clear, whether unstructured data sources can actually provide predictive information about the survival of patients. 
Comparisons of our model with DeepSurv \parencite{Katzman2016} and a vanilla VAE embedding imply that our method works similar well with common modeling choices (see Table~\ref{tab:ctdata-validation}), but also indicate that extraction of information is indeed difficult. 
\begin{table}[t]
	\centering
	\begin{tabular}{lcccccc}
		\toprule
		& \multicolumn{3}{c}{C-index~$(\uparrow)$}  & \multicolumn{3}{c}{IBS~$(\downarrow)$} \\
		\cmidrule(lr){2-4}\cmidrule(lr){5-7}
		& \multicolumn{1}{c}{Base} & \multicolumn{1}{c}{CoxPH} & 
		\multicolumn{1}{c}{RSF} & \multicolumn{1}{c}{Base} & \multicolumn{1}{c}{CoxPH} & 
		\multicolumn{1}{c}{RSF} \\ 
		\midrule
		CoxVAE &  \textbf{0.560} & 0.546 & 0.558 & 0.187 & 0.162 & 0.169\\ 
		VAE & -- & 0.554 & 0.544 & -- & 0.158 & \textbf{0.153} \\ 
		DeepSurv & 0.548 & \textbf{0.560} & 0.539 & 0.200 & 0.162 & 0.176 \\ 
		\bottomrule
	\end{tabular}
	\vspace{0.3cm}
	\caption{Validation set results for the CTs with liver metastases using the C-index \parencite{Harrell1982} and integrated Brier score (IBS; \cite{Graf1999}). The DeepSurv embedding is obtained by extracting the activations of the last hidden layer. We compute the respective embedding for each model and subsequently fit a Cox PH model and a random survival forest (RSF, \parencite{Ishwaran2008}) on the obtained latent variables as downstream task. The \textit{base} columns reflect the prediction performance of the architectures themselves.} 
	\label{tab:ctdata-validation}
\end{table}

In conclusion, we have demonstrated how a hazard-regularized variational autoencoder can be fitted to unstructured image data, thereby imprinting survival information into the latent space. This latent space is explainable and its relationship with the survival outcome can be easily visualized using the model's decoder. We demonstrated this approach exemplarily for survival data of cancer patients with metastases in the liver, where the generative model learned that increased risk is associated with increased tumor load.

\section*{Acknowledgments}

This work has been partially supported by the German Federal Ministry of Education and Research (BMBF) under Grant No. 01IS18036A.
We thank the anonymous reviewers for their constructive comments, which helped us to improve the manuscript.

\printbibliography

\end{document}